\documentclass[]{interact}
\usepackage{wrapfig}
\usepackage{color}
\usepackage[dvipsnames]{xcolor}

\usepackage{epstopdf}
\usepackage[caption=false]{subfig}

\usepackage[numbers,sort&compress]{natbib}
\bibpunct[, ]{[}{]}{,}{n}{,}{,}
\renewcommand\bibfont{\fontsize{10}{12}\selectfont}
\makeatletter
\def\NAT@def@citea{\def\@citea{\NAT@separator}}
\makeatother

\theoremstyle{plain}

\theoremstyle{definition}

\theoremstyle{remark}

\usepackage{multirow}
\usepackage{graphicx}
\usepackage{tabularx}

\usepackage{url} 

\begin{document}

\title{
Pre-Manipulation Alignment Prediction with \\Parallel Deep State-Space and Transformer Models
}

\author{
    \name{
        Motonari Kambara and Komei Sugiura
        \thanks{
            CONTACT Motonari Kambara. Email: motonari.k714@keio.jp
            \\\\This is a preprint of an article submitted for consideration in ADVANCED ROBOTICS, copyright Taylor \& Francis and Robotics Society of Japan; ADVANCED ROBOTICS is available online at http://www.tandfonline.com/.
        }
    }
    \affil{
        Keio University, 3-14-1 Hiyoshi, Kohoku, Yokohama, Kanagawa 223-8522, Japan
    }
}

\maketitle

\begin{abstract}
In this work, we address the problem of predicting the future success of open-vocabulary object manipulation tasks. 
Conventional approaches typically determine success or failure after the action has been carried out.
However, they make it difficult to prevent potential hazards and rely on failures to trigger replanning, thereby reducing the efficiency of object manipulation sequences.
To overcome these challenges, we propose a model, which predicts the alignment between a pre-manipulation egocentric image with the planned trajectory and a given natural language instruction.
We introduce a Multi-Level Trajectory Fusion module, which employs a state-of-the-art deep state-space model and a transformer encoder in parallel to capture multi-level time-series self-correlation within the end effector trajectory. 
Our experimental results indicate that the proposed method outperformed existing methods, including foundation models.
\end{abstract}



\section{Introduction \label{intro}}
\begin{figure}[h]
    \centering
    \includegraphics[width=\linewidth]{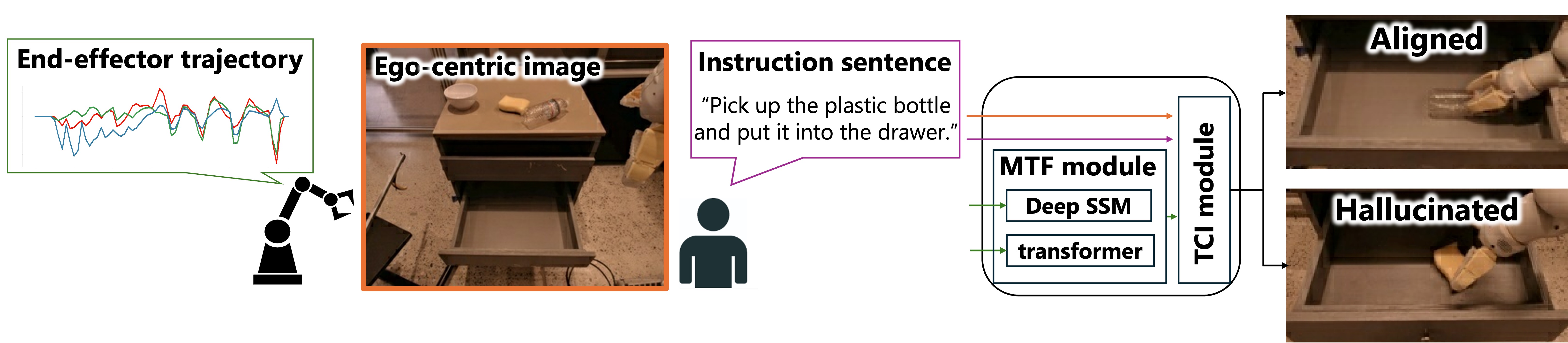}
    \vspace{-12pt}
    \caption{ \small
Overview of our method. In this figure, `Deep SSM', `MTF module', and `TCI module' represent a state-of-the-art deep state-space model, the Multi-Level Trajectory Fusion module, and the Trimodal Cross Fusion module, respectively. 
The input to the model consists of an egocentric image before the manipulation, a natural language instruction, and an end-effector trajectory. 
The model should output a prediction probability as to whether these inputs are aligned or hallucinated.}
\label{fig:task}
\end{figure}
Object manipulation is essential in various fields such as household tasks~\cite{wu2023tidybot} and agriculture~\cite{lehnert2017autonomous, jun2021towards}.
Insights into manipulation task success enhance efficiency and safety by preventing hazards stemming from failures and avoiding redundant tasks after object manipulation failures.

This study focuses on predicting that instruction sentences, egocentric images, and planned trajectories are aligned or hallucinated.
In this task, the planned trajectory incorporates information on the gripper's opening and closing states, which significantly increases the overall complexity. 
Consequently, this task is especially challenging, as it demands reliable prediction of future interactions between objects and the end-effector.
Such prediction relies on analyzing the trajectory and ensuring these interactions align with the natural language instructions.
A typical task example is illustrated in Fig.~\ref{fig:task}. 

This task is important because it has the potential to improve the overall efficiency of object manipulation.
Most existing methods \cite{goko2024task, xiao2022skill, liu2023reflect, inceoglu2021fino, driess2023palm, diehl2022did, das2021semantic} related to the prediction of manipulation success perform the actual success/failure prediction after the manipulation has been performed.
These approaches are fundamentally limited by their reliance on evaluating the result only after the manipulation is complete, making it difficult to mitigate risks during the manipulation.
Furthermore, replanning based on execution failures inherently limits the efficiency of robotic manipulation.
A more desirable approach would be to validate planned trajectories a priori by assessing their consistency with the task context, which includes both visual observations and natural language instructions.
This proactive validation potentially prevents costly execution errors and improves overall task completion.
While Vision-Language-Action (VLA) models~\cite{octo_2023, brohan2022rt, kim24openvla} are designed to address this limitation by performing alignment between visual observations, language instructions, and planned trajectories, they often fail to achieve sufficient alignment in practice.
Indeed, for instance, OpenVLA~\cite{kim24openvla}, a state-of-the-art VLA model, achieved almost 0\% success rate in zero-shot manipulation tasks~\cite{black2024pi_0}.

This study proposes a model for pre-manipulation alignment prediction between natural language instructions, egocentric images, and planned trajectories.
We introduce the Multi-Level Trajectory Fusion (MTF) module to obtain multi-level time-series information from trajectories. 
The MTF module performs parallel processing using two architectures: transformers (e.g.,~\cite{vaswani2017attention}), which are well-suited for capturing correlations in relatively short sequences, and state-of-the-art deep state-space models (e.g.,~\cite{s4}), which are reported to handle long sequences more effectively than transformers.
This alignment enables the interactions between objects and the end effector to be determined from images and trajectories. 
Consequently, unlike many existing methods, our approach enables predicting alignment between a natural language instruction, an image, and a planned trajectory before the manipulation execution. 
Furthermore, we introduce the Trimodal Cross-Integration module.  
This module applies cross-attention to compute correlations among three input modalities: natural language, images, and trajectories.  
By leveraging the inter-modal correlations, the model can consider whether a given trajectory executed in the environment aligns with the natural language instruction.

The main contributions of this research are as follows:
\begin{itemize}
\item We propose a model for verifying the pre-manipulation alignment between natural language instructions, a planned trajectory, and an egocentric image.
\item We introduce the MTF module, which employs a state-of-the-art state-space model and a transformer encoder in parallel to capture multi-level time-series self-correlation within the end effector trajectory.
\item We also introduce the Trimodal Cross-Integration module to effectively align natural language instructions, camera images, and the end effector trajectories.
\end{itemize}

\section{Related Work}
\subsection{Task Success Prediction}
Numerous studies have explored robot collision and discussed the approaches employed to maintain safety when humans and robots share a workspace~\cite{haddadin2017robot, mottaghi2016what, magassouba2021predicting, kambara2022relational}.
Several methods predict collisions using images and placement policies~\cite{mottaghi2016what, magassouba2021predicting, kambara2022relational}. 
Our method differs from these approaches by considering potential risks beyond collisions that may contribute to task failure.

Liu et al. proposed Failure Classifier~\cite{liu2024model} to detect erroneous trajectory generation based on images and predicted trajectories. 
However, their model cannot handle natural language instructions, making it unsuitable for direct application to the task addressed in this paper.

The proposed method is also closely related to subtask planning methods in long-horizon tasks~\cite{kawaharazuka2024real, brohan2023saycan, brohan2022rt}. 
Several approaches determine the success or failure of a task after executing a subtask and then replan subsequent subtasks based on that assessment using Vision Language Models~\cite{shirasaka2024selfrecovery, Driess2023palme, brohan2023saycan, zha2024distilling, sun2023interactive, zhang2024feedback, zhi2024closed, lin2024drplanner, xiong2024autonomous, sermanet2024robovqa, duan24aha}. 
These methods perform incremental success/failure evaluation and re-planning for each subtask in the process of object manipulation. 
In particular, REFLECT~\cite{liu2023robofail} verifies whether predefined states are achieved based on the target states associated with each object class. 
Contrastive $\lambda$-Repformer~\cite{goko2024task} is similar to our proposed approach in that it can be considered as a predictive model of success or failure in open-vocabulary object manipulation tasks. 
However, because these existing methods are assumed to be applied post-manipulation, they inherently rely on an evaluation of the outcome after the completion of object manipulation. 
As a result, it is challenging to prevent hazards during manipulation, such as an object slipping or dropping while being grasped. 
Furthermore, replanning requires failure before it can be initiated, reducing the efficiency of object manipulation sequences.
In contrast, our method predicts future interactions with the environment based on the end effector trajectory.
This enables the prediction of the pre-manipulation alignment, which is indicative of the potential for successful manipulation.

\subsection{Visual Question Answering (VQA)}
Our task is closely related to the VQA task~\cite{antol2015vqa}, which takes an image and a natural-language question as input and outputs an appropriate text-based answer. 
In this study, we show that the task we address can also be handled by VQA models if a suitable prompt is provided, including both the manipulation instruction and the camera image taken prior to object manipulation with the end-effector trajectory is overlaid (see Sec.~\ref{baseline}).

Extensive work has focused on VQA tasks~\cite{yin2023survey, wu2017visual}. In particular, foundation models, which have been rapidly improving via large-scale pretraining on multiple tasks, have demonstrated remarkable performance in VQA~\cite{wang2024qwen2, li2023blip, bai2023qwen, wang2023image}. Indeed, the effectiveness of instruction tuning, which is widely regarded to enhance the performance of foundation models, has been validated for VQA tasks~\cite{liu2024improved, liu2024visual, instructblip}. Furthermore, numerous standard benchmarks have been proposed for VQA tasks, facilitating performance comparisons among these methods~\cite{hudson2019gqa, goyal2017making}.

Existing VQA models exhibit sophisticated spatial understanding, reasoning capabilities, and the ability to follow instructions, thereby achieving favorable performance across a wide range of domains~\cite{wang2024qwen2}. 
However, as shown in Sec.~\ref{results}, even state-of-the-art VQA models find our task challenging, likely because it requires more advanced spatial reasoning for the appropriate alignment of the trajectory with the image. 
To address this, our proposed method introduces a cross-attention mechanism that explicitly accounts for the alignment between the trajectory and the image.
\section{Problem Statement}
This study focuses on Pre-Manipulation Alignment Prediction (PMAP) for open-vocabulary object manipulation.
This task involves verifying the alignment between a natural language instruction sentence, a pre-manipulation egocentric image, and a planned trajectory.
Given the pre-manipulation image, the trajectory, and the instruction sentence, the model should determine whether they are mutually aligned, indicating a likely successful manipulation.
Fig.~\ref{fig:task} illustrates a typical example of the PMAP task. 
In this episode, the natural language instruction is given as ``Pick up the plastic bottle and put it into the drawer.''
In this example, models should predict `Aligned' if the manipulator puts a plastic bottle into a drawer, and `Hallucinated' if it places a sponge.

Handling natural language instructions is crucial in this task for the following reasons. 
In scenarios where the task is not pre-defined but instead specified via natural language, the success or failure of an object manipulation can vary depending on the instruction. 
Therefore, it is essential that the robot properly interprets the given language instruction.

For instance, as shown in Fig. 1, if the task were ``Put the sponge into the drawer,'' the manipulation would likely be considered to be successful. 
However, because the actual instruction is ``Pick up the plastic bottle and put it in the drawer,'' the appropriate prediction is that the manipulation will likely fail.

In the PMAP task, the input consists of an egocentric image, the trajectory of the end effector, and a natural language instruction sentence.
The expected output is the predicted probability $p(\hat{y}=1)$. 
$y$ and $\hat{y}$ denote a label and predicted label, respectively. 
The condition $y=1$ indicates that the planned trajectory, the egocentric image, and the given natural language instruction are aligned. 
The input image is exclusively the one captured just before the manipulation. 
Consequently, the model is required to infer the final state of the manipulation from the trajectory data and this single pre-manipulation image, making the task particularly challenging.
In this study, we assume that trajectory generation is out of scope.

\section{Proposed Method}

\begin{figure}[t]
    \centering
    \includegraphics[width=\linewidth]{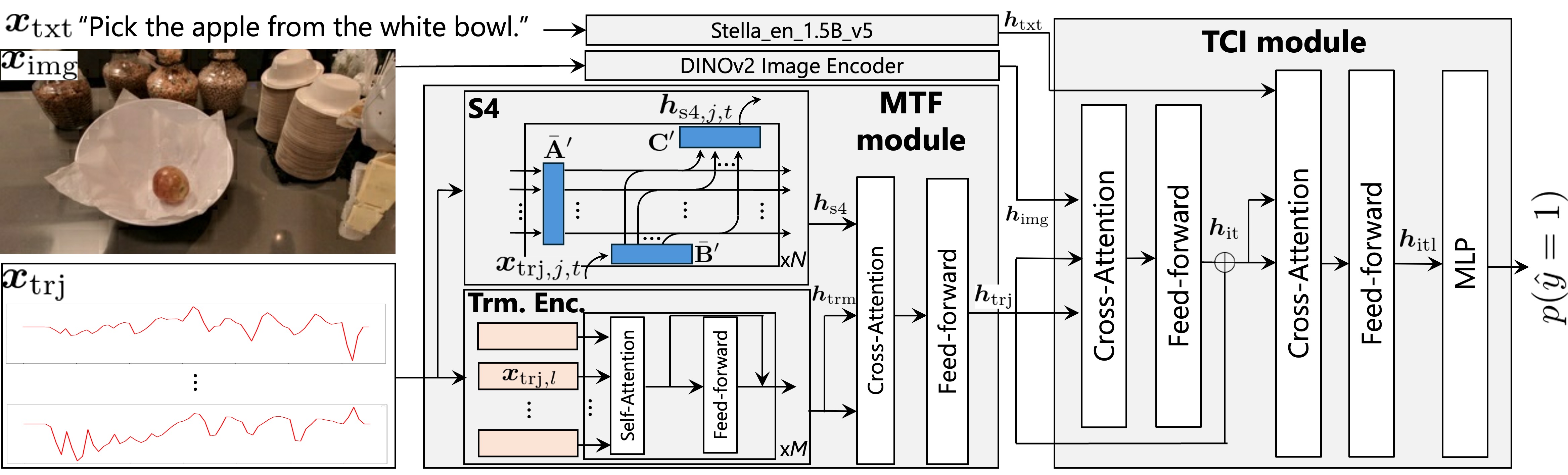}
    \vspace{-8pt}
    \caption{
    \small Network overview of the proposed method. In this figure, `MTF module', `TCI module', $N$, and $M$ denote the Multi-Level Trajectory Fusion module, the Trimodal Cross-Fusion module, the number of S4 blocks, and the number of transformer layers, respectively. Furthermore, $l$ and $j$ correspond to the index along the temporal dimension and the index of the degrees of freedom of the end effector, respectively. We omit $D$ in Eq. (6) to improve the readability. The variables $\bar{\bm{A}}'$, $\bar{\bm{B}}'$, and $\bm{C}'$ correspond to the matrices defined in Eq.~(6).}
    \label{fig:model}
\end{figure}

Fig. \ref{fig:model} shows an overview of the proposed method, which consists of the Multi-Level Trajectory Fusion (MTF) module and the Trimodal Cross-Integration (TCI) module. 
The main novelties of our method are as follows:
\begin{itemize}
    \item We propose a model that predicts the alignment between language instructions, pre-manipulation images, and trajectories.
    \item We introduce the MTF module, which captures temporal relationships in trajectories using parallel deep SSM and transformer encoders.
    \item We introduce the TCI module to acquire a shared representation between the input modalities.
\end{itemize}

\subsection{Input}
The input to the model, denoted as \(\bm{x}\), is defined as:
\begin{align}
    \bm{x} &= \left\{\bm{x}_\mathrm{txt}, \bm{x}_\mathrm{img}, \bm{x}_\mathrm{trj}\right\},
\end{align}
where \(\bm{x}_\mathrm{txt}\), \(\bm{x}_\mathrm{img}\), and $\bm{x}_\mathrm{trj} \in \mathbb{R}^{J \times T}$ represent the natural language instruction, camera image, and end effector trajectory, respectively. Here, $J$ and \(T\) denote the number of degrees of freedom of the manipulator and the length of the time series, respectively.

\subsection{Multi-Level Trajectory Fusion module}
We introduce the Multi-Level Trajectory Fusion (MTF) module, a module designed to filter the end effector trajectory using learnable weights. This filtering process enables temporal trajectory compression for more efficient representation and processing.

The input to this module is $\bm{x}_{\mathrm{trj}}$, and the output is $\bm{h}_{\mathrm{trj}}$.
This module is composed of a transformer encoder, a Structured State Space Sequence Model (S4) module~\cite{s4}, and a cross-attention mechanism. 
In the field of robotics, transformer-based approaches are frequently used to capture time-series dependencies in trajectories (e.g., \cite{brohan2023saycan}). 
On the other hand, for long sequence signals, deep state-space models (deep SSMs) outperform transformers in terms of capturing temporal relationships~\cite{s4}. 
The length of the end effector trajectory is determined by the product of its control frequency and the overall execution time. 
More complex object manipulations generally take longer to complete, resulting trajectory tends to be longer. 
Therefore, as shown in Fig. \ref{fig:model}, this module applies a transformer to each sub-trajectory while employing a deep SSM over the entire trajectory, thereby extracting time-series dependencies at multiple levels.

\subsubsection{Structured State Space Sequence Model (S4) \label{s4_exp}}
Deep SSMs \cite{lssl, s4, s4d, s5} have recently emerged as a powerful approach for long sequence modeling, demonstrating superior performance over existing architectures, including transformers, on various benchmarks~\cite{s4}. 
These models, while incorporating concepts from classical SSM theory \cite{classic_ssm, classic_ssm2}, leverage advancements inspired by RNNs and CNNs to effectively capture dependencies in long sequence data.
Several deep SSMs, including S4~\cite{s4} and Mamba~\cite{gu2024mamba}, are inspired by continuous systems that map 1-D input signals $x_\mathrm{in}(t) \in \mathbb{R}$~\cite{zhuvision}. In this continuous system, an input signal $x_\mathrm{in}(t) \in \mathbb{R}$ is mapped to an output signal $y_\mathrm{out}(t) \in \mathbb{R}$ via a hidden state $\bm{h}(t) \in \mathbb{R}^{H \times 1}$, where $H$ is the dimensionality of the hidden representation. This mapping is defined as follows:
\begin{equation}
\begin{aligned}
    \label{eq:continuous_ssm}
    \bm{h}'(t) &= \mathbf{A}\,\bm{h}(t) + \mathbf{B}\,x_\mathrm{in}(t),\\
    y_\mathrm{out}(t) &= \mathbf{C}\,\bm{h}(t) + D\,x_\mathrm{in}(t),
\end{aligned}
\end{equation}
where $\mathbf{A} \in \mathbb{R}^{H \times H}$, $\mathbf{B} \in \mathbb{R}^{H \times 1}$, $\mathbf{C} \in \mathbb{R}^{1 \times H}$, and $D \in \mathbb{R}$ are learnable parameters.

To process sequences in discrete time steps, a discretization method such as Zero-Order Hold \cite{zoh} is typically employed~\cite{zhuvision}, assuming the input $x_\mathrm{in}(t)$ is held constant over the step size for discretization $\Delta=[t_{k-1},t_k]$. With Zero-Order Hold, we obtain:
\begin{equation}
\begin{aligned}
    \label{s4block}
    \bm{h}_k &= \bar{\mathbf{A}}\,\bm{h}_{k-1} + \bar{\mathbf{B}}\,x_{\mathrm{in},k},\\
    y_{\mathrm{out}, k} &= \mathbf{C}\,\bm{h}_k + D\,x_{\mathrm{in},k},
\end{aligned}
\end{equation}
where $\bar{\mathbf{A}}=\exp(\Delta \mathbf{A})$ and $\bar{\mathbf{B}}=(\Delta \mathbf{A})^{-1}\bigl(\exp(\Delta \mathbf{A})-\mathbf{I}\bigr)\Delta \mathbf{B}$. While this recurrent-style formulation captures self-correlation, it is generally not easy to apply parallelization across time steps.

S4 \cite{s4} resolves this limitation by re-expressing the discrete SSM as a convolution. Specifically, it constructs a structured kernel,
\begin{align}
    \bar{\mathbf{K}} = \bigl(D + \mathbf{C}\bar{\mathbf{B}},\,D +\mathbf{C}\bar{\mathbf{A}}\bar{\mathbf{B}},\,\ldots,\,D +\mathbf{C}\bar{\mathbf{A}}^{S-1}\bar{\mathbf{B}}\bigr),
\end{align}
and computes the output via
\begin{align}
    \bm{y}_\mathrm{out} = \bm{x}_\mathrm{in} \ast \bar{\mathbf{K}},
\end{align}
where $S$ represents the sequence length of $\bm{x}_\mathrm{in}$. This approach allows training to be parallelized across the entire sequence, while also permitting a recurrent-style update for efficient inference. By unifying the state-space dynamics with deep neural network techniques, S4 achieves convolutional parallelism during training and efficient recurrence during inference, making it highly suitable for large-scale sequence tasks.

When processing multi-channel input $\mathbf{X}_\mathrm{in}=\{\bm{x}'_{\mathrm{in},k} \in \mathbb{R}^{N_c \times 1}|k=1,...,S\}$, $N_c$ independent S4 blocks are used, incorporating channel mixing to capture inter-channel correlations. 
Specifically, as described in \cite{s5}, Eq.~\ref{s4block} is extended as follows:
\begin{equation}
\begin{aligned}
    \label{s4block_multi}
    \mathbf{H}_k &= \bar{\mathbf{A}}'\,\mathbf{H}_{k-1} + \bar{\mathbf{B}}'\,\bm{x}'_{\mathrm{in},k},\\
    \bm{y}'_{\mathrm{out}, k} &= \mathbf{C}'\,\mathbf{H}_k + \mathbf{D}'\,\bm{x}'_{\mathrm{in},k},
\end{aligned}
\end{equation}
where $\mathbf{H}_{k}=[\bm{h}_{k,1}^\top, \bm{h}_{k,2}^\top,...,\bm{h}_{k,N_c}^\top]^\top$, with $\bm{h}_{k,v} \in \mathbb{R}^{1 \times H}$. Here, $\bar{\mathbf{A}}' \in \mathbb{R}^{HN_c \times HN_c}$, $\bar{\mathbf{B}}' \in \mathbb{R}^{HN_c \times N_c}$, $\mathbf{C}' \in \mathbb{R}^{N_c \times HN_c}$, and $\mathbf{D}' \in \mathbb{R}^{N_c \times N_c}$.
In each S4 layer, a linear transformation is ultimately performed on $\bm{y}'_{\mathrm{out}, k}$ for mixing.

\subsubsection{Module Architecture}
We first provide an explanation of the S4 module. 

The S4 module consists of the $N$ S4 blocks. As described in Sec. 4.2.1, each S4 block for the multi-channel input can be defined by Eq. 6. 
In this module, the variable $\bm{x}'_{\mathrm{in},k}$ and $\bm{y}'_{\mathrm{out},k}$ in Eq. 6 are replaced by $\bm{x}_{\mathrm{trj}, j, t} \quad (j \in \{1,\dots,J\}, t \in\{1,...,T\})$ and $\bm{h}_{\mathrm{s4},j,t}$, respectively. 
Finally, we apply average pooling and linear transformation, and obtain an intermediate feature $\bm{h}_{\mathrm{s4}} \in \mathbb{R}^{d'}$, where $d'$ denotes the dimensionality of the weights.

Next, we describe how the transformer encoder processes the input. The transformer encoder has the $M$ transformer layer. The trajectory $\bm{x}_{\mathrm{trj}}$ is split after every $L$ steps to form $\bm{x}'_{\mathrm{trj}} = \{\bm{x}_{\mathrm{trj},l} \in \mathbb{R}^{L \times J} \mid l=1,\dots, L\}.$ 
By feeding this token sequence into the transformer encoder, we obtain an intermediate feature $\bm{h}_{\mathrm{trm}} \in \mathbb{R}^{d'}$: 
\begin{align}
    \bm{h}_{\mathrm{trm}} &= f_{\mathrm{trm}}\bigl(\bm{x}'_{\mathrm{trj}}\bigr),
\end{align}
where $f_{\mathrm{trm}}(\cdot)$ represents the transformer encoder. Finally, the cross-attention between $\bm{h}_{\mathrm{s4}}$ and $\bm{h}_{\mathrm{trm}}$ is computed. The vector $\bm{h}_{\mathrm{trj}} \in \mathbb{R}^{d'}$ is the final output of this module.

\subsection{Trimodal Cross-Integration module}
The PMAP task requires models to predict the alignment between a natural language instruction, an image, and a planned trajectory.
To achieve this alignment prediction, an understanding of each object's positional information within the input image is necessary to determine whether the planned trajectory corresponds to the described manipulation.

Therefore, we introduce the Trimodal Cross-Integration (TCI) module. 
By computing the alignment among the natural language instructions, trajectory, and image information, this module generates the probability of whether the end effector trajectory in the observed environment will successfully accomplish the task specified by the natural language instructions.

The TCI module consists of two transformer decoders and an MLP layer. The input to this module is 
$\bm{h}_\mathrm{txt}, \bm{h}_\mathrm{img}, \bm{h}_\mathrm{trj}$ 
and the output is the predicted probability $p(\hat{y})$ of $\bm{x}_{\mathrm{txt}}$, $\bm{x}_{\mathrm{img}}$, and $\bm{x}_{\mathrm{trj}}$ are aligned.
Here, $\bm{h}_\mathrm{txt}$ and $\bm{h}_\mathrm{img}$ are the language features obtained by embedding $\bm{x}_\mathrm{txt}$ using Stella\_en\_1.5B\_v5~\cite{stellaen} and the image features extracted from $\bm{x}_\mathrm{img}$ using DINOv2~\cite{oquab2024dinov2}, respectively.

First, we compute the cross-attention between $\bm{h}_\mathrm{img}$ and $\bm{h}_\mathrm{trj}$ as follows:
\begin{align}
\label{eq:cross_attention_img_traj}
\bm{h}_\mathrm{it} &= \text{CrossAttn}\bigl(\bm{h}_\mathrm{trj}, \bm{h}_\mathrm{img}\bigr),
\end{align}
where $\text{CrossAttn}\bigl(\bm{X}_a, \bm{X}_b\bigr)$ is defined as:
\begin{align}
\text{CrossAttn}\bigl(\bm{X}_a, \bm{X}_b\bigr) 
&= \text{softmax}\!\left(\frac{\bm{X}_a \bm{W}_q (\bm{X}_b \bm{W}_k)^\top}{\sqrt{d_k}}\right)\bm{X}_b \bm{W}_v,
\end{align}
where $\bm{W}_q, \bm{W}_k, \bm{W}_v \in \mathbb{R}^{d_\text{in} \times d_\text{out}}$ are trainable weights, and $d_k$ is the dimensionality of $\bm{X}_b \bm{W}_k$. 
The resulting $\bm{h}_\mathrm{it}$ represents aligned features between the image and trajectory modalities.
Next, we compute the cross-attention between $\bm{h}_\mathrm{txt}$ and $\bm{h}_\mathrm{it}$:
\begin{align}
\bm{h}_\mathrm{itl} &= \text{CrossAttn}\bigl(\bm{h}_\mathrm{txt}, \bm{h}_\mathrm{it}\bigr),
\end{align}
where $\bm{h}_\mathrm{itl}$ is an aligned feature between $\bm{h}_\mathrm{it}$ and $\bm{h}_\mathrm{txt}$, reflecting whether the natural language information related to object manipulation aligns with the environment information conditioned on trajectories.
Finally, we obtain $p(\hat{y})$ from $\bm{h}_{itl}$ using a fully connected layer as follows:
\begin{align}
p(\hat{y}) &= \text{softmax}\bigl(\bm{W}_o \bm{h}_\mathrm{itl} + \bm{b}_o\bigr),
\end{align}
where $\bm{W}_o$ and $\bm{b}_o$ are trainable parameters.

\section{Experimental Settings}
\begin{table}[t]
    \normalsize
    \centering
    \caption{\small Experimental settings of the proposed method.}
    \label{tab:params}
    \begin{tabular}{lc}
        \toprule
        $M$ & $2$ \\
        $N$ & $4$ \\
        $d'$ & $512$ \\
        Optimizer & Adam ($\beta_1=0.9$, $\beta_2=0.98$) \\
        Learning rate & $1\times10^{-5}$ \\
        Batch size & $64$ \\
        \#Epoch & $50$ \\
        \bottomrule
        \end{tabular}
\end{table}

\subsection{Dataset}
We constructed a dataset named SP-RT-1-Traj, by extending the SP-RT-1 dataset~\cite{goko2024task}.
The SP-RT-1 dataset was built based on the RT-1 dataset~\cite{brohan2022rt}. The SP-RT-1 dataset includes images before and after object manipulation, object manipulation instruction sentences, and success/failure labels for each episode of object manipulation. 
In contrast, the PMAP task requires a dataset that includes egocentric images before object manipulation, end effector trajectories, natural language instructions, and task outcome labels indicating success or failure.
Therefore, because the SP-RT-1 dataset generally contains the information required for the PMAP task, we constructed our SP-RT-1-Traj dataset using the SP-RT-1 dataset.

The end effector trajectories in each episode were found to be lacking in the SP-RT-1 dataset. Therefore, we collected the trajectories from the RT-1 dataset for the PMAP task.
We split the dataset following the description in~\cite{goko2024task}.
Specifically, the test set of SP-RT-1-Traj shares the same episodes as the test split of the original SP-RT-1 dataset. 
As reported by Goko et al., these splits were constructed by randomly splitting the entire data set.

The constructed dataset includes 13,915 samples, split into 11,915 training samples, 1,000 validation samples, and 1,000 test samples.
The manipulator used for data collection had a seven-dimensional action space~\cite{brohan2022rt}. 
An additional dimension was related to the opening and closing of the end effector.
Therefore, the trajectory at each timestep has eight dimensions.

\subsection{Method Settings}
Table~\ref{tab:params} shows the experimental settings of the proposed method.
We used an NVIDIA GeForce RTX 4090 with 24GB of VRAM, 64GB of RAM, and an Intel Core i9--13900KF. The training time for the proposed method was approximately 0.6 hours, and the inference time, excluding feature extraction, was 0.34 ms per sample.

\subsection{Baseline Methods \label{baseline}}
\begin{figure}[ht]
    \centering
    \includegraphics[width=\linewidth]{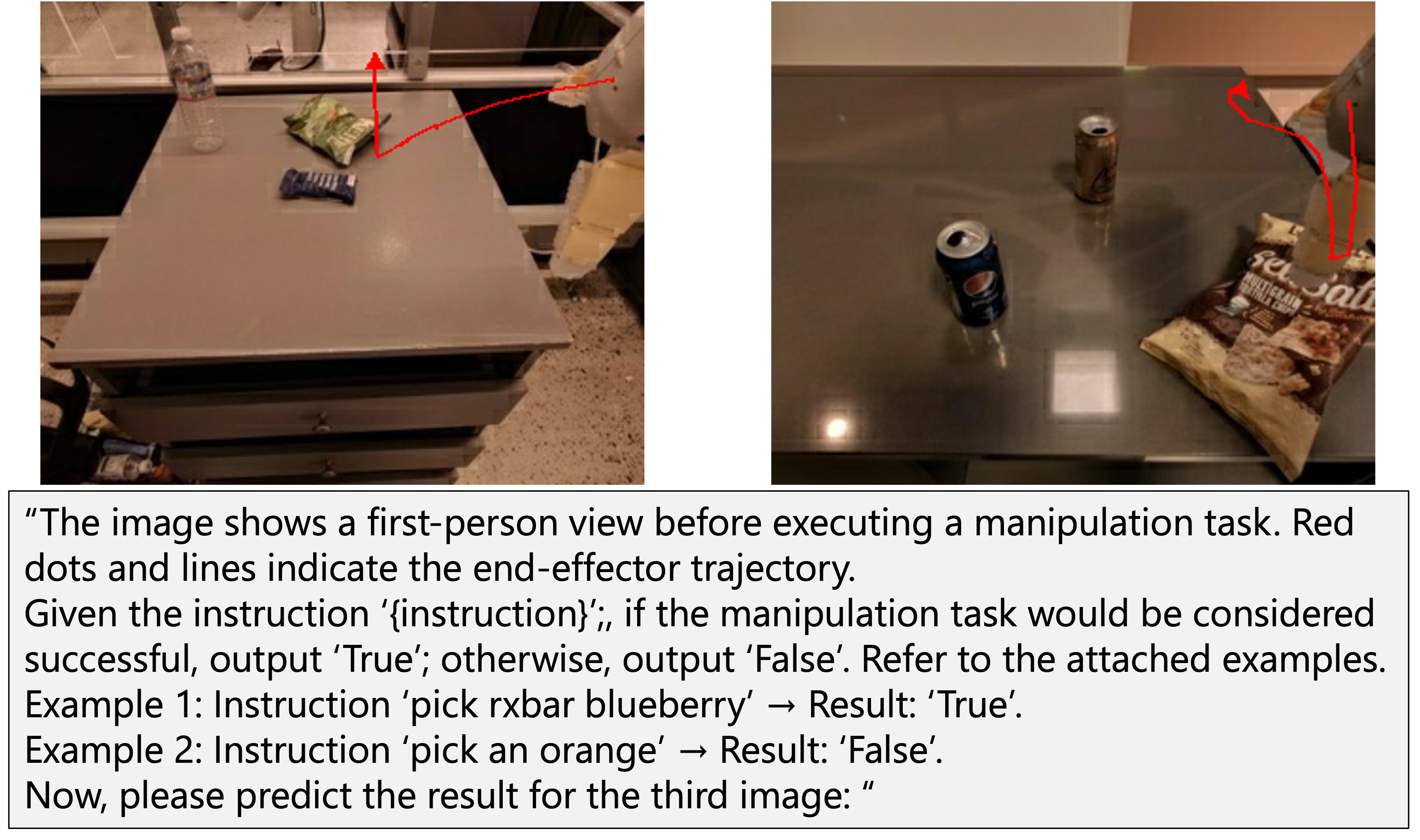}
    \vspace{-20pt}
    \caption{
    \small 
    The prompt input to GPT-4o. 
    The images on the left and right show the images used as prompts for Example 1 and 2.
    The red arrows in the images represent the trajectory of the end-effector. 
    }
\label{fig:prompt}
\end{figure}
In our experiments, we used GPT-4o~\cite{gpt4o}, Qwen2-VL~\cite{wang2024qwen2}, and Contrastive $\lambda$-Repformer~\cite{goko2024task} as the baseline methods. 
We selected these baseline methods for the following reasons:
GPT-4o is a representative multimodal large language model that has been pre-trained on large-scale datasets and has exhibited remarkable performance in a variety of tasks, including VQA. 
Additionally, Qwen2-VL is a state-of-the-art VQA model.
We chose Contrastive $\lambda$-Repformer because it has demonstrated remarkable results on the Success Prediction for Open-vocabulary Manipulation (SPOM) task~\cite{goko2024task}, which is closely related to the PMAP task. 

Contrastive $\lambda$-Repformer was trained on the SP-RT-1-Traj dataset, whereas experiments for other baseline models were conducted in zero/few-shot settings.
Post-manipulation images were not provided to Contrastive $\lambda$-Repformer to prevent leakage of the object manipulation outcomes. 
These baselines cannot directly accept trajectories as inputs; therefore, as illustrated in Fig.~\ref{fig:prompt}, we used images with the trajectory overlaid onto the pre-manipulation image as the input. 
Indeed, some studies employ vectorized trajectories for multimodal large language models input~\cite{kuangram, bahl2023affordances, hwang2025motif}.
%
The prompt given to GPT-4o is shown in Fig.~\ref{fig:prompt}. 
As shown in the figure, two samples were included in the prompt as examples. That is, GPT-4o was used in a few-shot setting.
The prompt given to Qwen2-VL was as follows:
``
The image shows a first-person view before executing a manipulation task.
Red dots and lines indicate the end-effector's trajectory.
Given the instruction `{description}', if the manipulation task would be considered successful, output `True'; otherwise, output `False'.
''
Since Qwen2-VL has difficulty handling multiple images as input, no sample examples were included in the prompt.

\section{Experimental Results}
\begin{table*}
    \centering
    \caption{ \small
        Quantitative results on the SP-RT-1-Traj dataset. Bold entries indicate the highest accuracy. 
    }
    {
    \begin{tabular}{lc}
        \toprule
        Method & Accuracy [\%] \\
        \hline
        Qwen2-VL~\cite{wang2024qwen2} & 52.1 $\pm$ 0 \\
        GPT-4o~\cite{gpt4o} & 69.4 $\pm$ 0.64 \\
        Contrastive $\lambda$-Repformer~\cite{goko2024task} & 77.7 $\pm$ 0.65 \\
        Ours & $\bm{80.1}$ $\pm$ 0.84 \\
        \hline
    \end{tabular}}
    \label{tab:quantitative}
\end{table*}

\subsection{Quantitative Results \label{results}}
Table~\ref{tab:quantitative} presents the quantitative results for comparing the baseline and proposed methods on the SP-RT-1-Traj dataset. Each score represents the mean and standard deviation. We conducted the experiments five times. 
The evaluation metric was accuracy.

The baseline methods Qwen2-VL, GPT-4o, and Contrastive $\lambda$-Repformer achieved accuracies of 52.1\%, 69.4\%, and 77.7\%, respectively. 
In contrast, our proposed method attained an accuracy of 80.1\%, outperformed the best baseline method, Contrastive $\lambda$-Repformer, by 2.4 percentage points. 
Furthermore, the performance differences were statistically significant $(p < 0.01)$. 
We attribute these results to the proposed method’s ability to predict future situations based on trajectory information more effectively than the other methods.

\subsection{Qualitative Results}
\begin{figure}[t]
    \centering
    \includegraphics[width=\linewidth]{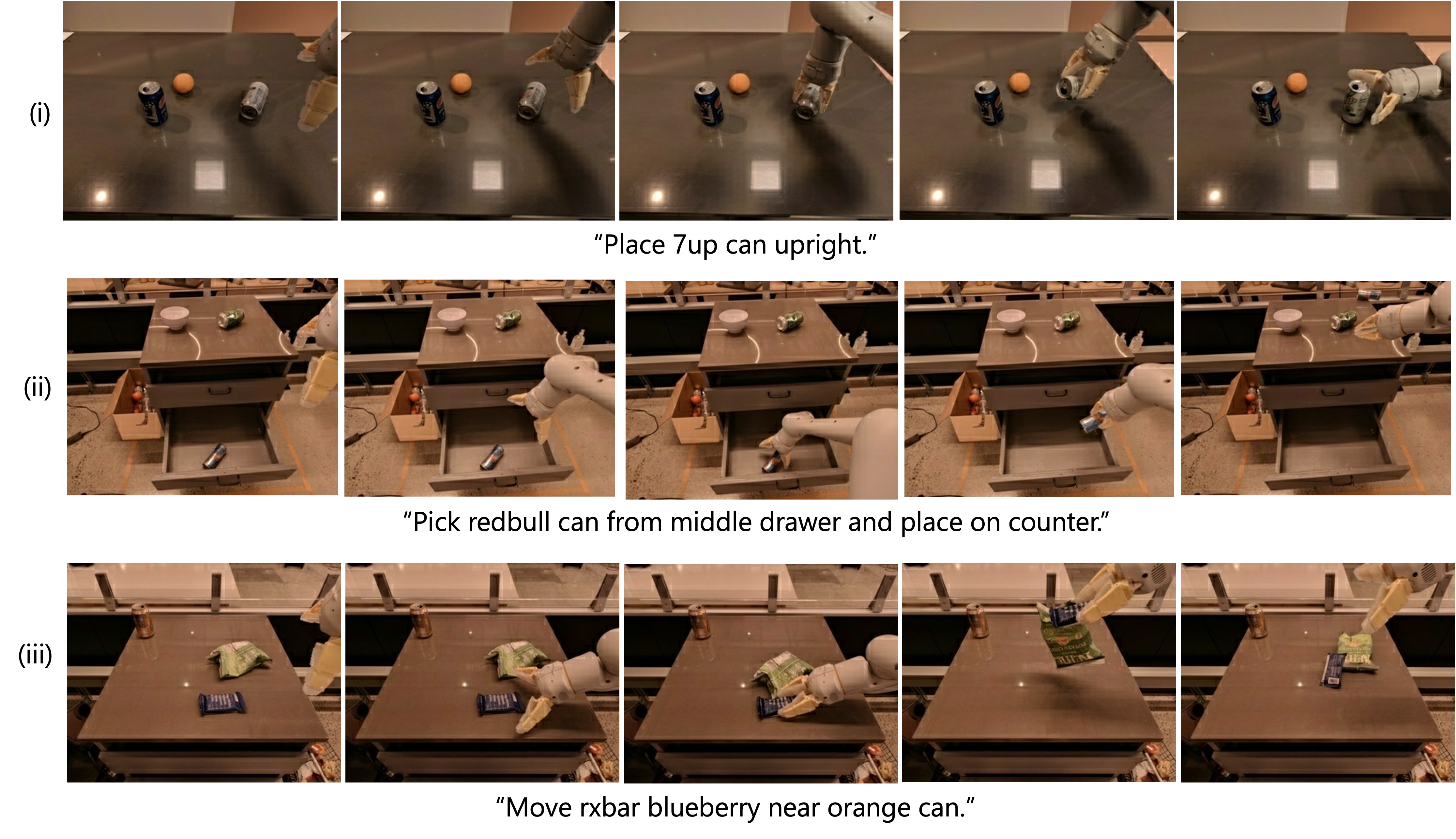}
    \vspace{-8pt}
    \caption{
    \small Qualitative results. Panels (i), (ii), and (iii) represent True Positive, True Negative, and False Positive examples, respectively. The leftmost image in each panel illustrates the scene before manipulation.
    }
    \label{fig:qualitative}
\end{figure}
Fig.~\ref{fig:qualitative} presents some qualitative results. 
Panels (i), (ii), and (iii) show examples of a True Positive, True Negative, and False Positive, respectively.
Each panel in Fig.~\ref{fig:qualitative} illustrates a sequence of images extracted from a specific episode, with the leftmost image captured prior to any object manipulation. 
Only this leftmost image was used as input to the model.

In the episode illustrated in Fig.~\ref{fig:qualitative} (i), the given instruction was ``Place 7up can upright.'' 
The environment contained two cans, and the manipulator selected the correct one and placed it upright without error. 
Thus, the inputs are appropriately aligned.
While the proposed method correctly predicted this sample, all baseline methods incorrectly predicted this. 
This result suggests that, in our proposed method, the TCI module effectively aligns text and visual data, thereby allowing the model to identify which can is the intended target and appropriately recognize its state.

In Fig.~\ref{fig:qualitative} (ii), the instruction was ``Pick redbull can from middle drawer and place on counter.'' 
The manipulator successfully grasped the can from inside the drawer and moved it above the counter. 
However, the suboptimal grasp position allowed the can to slip and roll away before being placed. 
That is, the trajectory was considered to be hallucinated.
The proposed method correctly predicted this sample, whereas GPT-4o and Contrastive $\lambda$-Repformer made incorrect inferences.
This indicates that our method leverages trajectory information to give a more reliable prediction of whether the planned trajectory and the egocentric image are aligned for the instruction sentence.

In contrast, Fig.~\ref{fig:qualitative} (iii) presents a case where our method produced an incorrect prediction. 
The instruction was ``Move rxbar blueberry near orange can.'' 
Although the manipulator grasped the blueberry bar and placed it closer to the orange can than its original location, the distance between the bar and the can still could not be considered sufficiently close.
Hence, the label was `Hallucinated'.
The proposed method predicted this sample incorrectly.
This episode required the spatial relationships among multiple objects to be recognized, therefore, correctly predicting the alignment was challenging.

\subsection{Ablation Study}
\begin{table*}
    \centering
    \caption{
    \small
    Ablation study on the performance of the MTF module. In this table, `Trm.' denotes the transformer encoder.
    }
    {
    \begin{tabular}{lcccc}
        \toprule
        Model & S4~\cite{s4} & Trm. & Accuracy [\%] \\
        \midrule
        (i) &  $\checkmark$ & $\checkmark$ & \textbf{80.1} $\pm$ 0.84 \\
        (ii) & $\checkmark$ &  & 79.7 $\pm$ 1.15 \\
        (iii) &  & $\checkmark$ & 79.9 $\pm$ 0.78 \\
        (iv) &  &  & 71.9 $\pm$ 1.76 \\
        \bottomrule \\
    \end{tabular}
    \vspace{-15pt}
    }
    \label{tab:te_ablation}
\end{table*}

\begin{table*}    \centering
    \caption{
    \small Ablation study on the performance of the cross-attention mechanism between trajectory and image features in the TCI module. The check mark in the `Cross-Attention' column indicates whether the procedure described in Eq.~\eqref{eq:cross_attention_img_traj} is applied.}
    {
    \begin{tabular}{lcc}
        \toprule
        Model & Cross-Attention & Accuracy [\%] \\
        \midrule
        (i) & $\checkmark$ & \textbf{80.1} $\pm$ 0.84 \\
        (ii) &  & 75.4 $\pm$ 1.02 \\
        \bottomrule \\
    \end{tabular}
    \vspace{-15pt}
    }
    \label{tab:tcfd_ablation}
\end{table*}

\begin{table*}[t]
\centering
\caption{\small Ablation study on the performance of modifying the feature pair combinations for cross-attention computation in the TCI module. The parentheses indicate which features were used in Eqs. (8) and (9).}
\begin{tabular}{lccc}
\toprule
Model & $\bm{h}_{\mathrm{it}}$ & $\bm{h}_{\mathrm{itl}}$ & Accuracy [\%] \\
\midrule
(i)   & ($\bm{h}_{\mathrm{txt}}, \bm{h}_{\mathrm{img}}$) & ($\bm{h}_{\mathrm{trj}}, \bm{h}_{\mathrm{it}}$) & 67.2 \\
(ii)  & ($\bm{h}_{\mathrm{trj}}, \bm{h}_{\mathrm{txt}}$) & ($\bm{h}_{\mathrm{img}}, \bm{h}_{\mathrm{it}}$) & 74.3 \\
(iii) & ($\bm{h}_{\mathrm{trj}}, \bm{h}_{\mathrm{img}}$) & ($\bm{h}_{\mathrm{txt}}, \bm{h}_{\mathrm{it}}$) & $\bm{80.1}$ \\
\bottomrule
\end{tabular}
    \vspace{-15pt}
    \label{tab:ca_ablation}
\end{table*}
\textbf{MTF module ablation:}
To explore how the S4 and transformer encoder architectures in the MTF module affect performance, we removed each component in turn.
The results are presented in Table~\ref{tab:te_ablation}, where Model~(iv) applies a fully connected layer to weigh the trajectories.
According to Table~\ref{tab:te_ablation}, the accuracy for Models~(ii), (iii), and (iv) decreases by 0.4, 0.2, and 8.2 points, respectively, compared with Model~(i).
These findings indicate that both S4 and the transformer encoder yield modest performance improvements.
Although there is no significant discrepancy in performance between S4 and the transformer encoder, removing either leads to a noticeable drop in accuracy.
This suggests that their respective weighting mechanisms are more effective than a fully connected layer in terms of extracting beneficial information.

\textbf{TCI module ablation:}
To assess how the cross-attention mechanism contributes to performance, we removed it from the model.
Table~\ref{tab:tcfd_ablation} demonstrates that the accuracy of Model~(ii) on the SP-RT-1-Traj dataset drops by 4.7 points in comparison to Model~(i).
This result suggests that pre-aligning image features with trajectory information enhanced the model's ability to comprehend the environment and the interactions within it.

\textbf{Cross-attention ablation:} To examine the effect of the ordering of feature pairs used for cross-attention in the TCI module—specifically when computing $\bm{h}_\mathrm{it}$ and $\bm{h}_\mathrm{itl}$—we varied the combinations. Table~\ref{tab:ca_ablation} shows the results. Compared to Model (iii), Models (i) and (ii) resulted in a performance decrease of 12.9 and 5.8 points, respectively.

These findings suggest that the configuration adopted in Model (iii), where cross-attention is first applied between the trajectory and the image (environment information), followed by attention with respect to the language instruction, yields better performance.

This observation can be attributed to the characteristics of object manipulation tasks, where execution is primarily influenced by the trajectory and the surrounding environment. Therefore, in the PMAP task, it is important to evaluate whether the trajectory-image correlation is consistent with the given language instruction.

\subsection{Error Analysis \label{error}}

\begin{table}[t]
    \caption{\small Categorization of failure cases. We randomly selected a total of 100 samples and performed a detailed manual analysis of the errors.}
    \label{tab:error}
    \normalsize
    \centering
    \begin{tabular}{lcc}
        \toprule
        Error Type & \# of Errors \\
        \midrule
        Object Grounding Error &
        42 \\
        Action Comprehension Error &
        27 \\
        Referring Expression Comprehension Error &
        23 \\
        Partial Visibility &
        7 \\
        Annotation Error &
        1 \\
        \midrule
        Total &
        100 \\
        \bottomrule
    \end{tabular}
\end{table}
The proposed method achieved the following results on the test set of the SP-RT-1-Traj dataset: 421, 383, 117, and 79 samples were classified as True Positive, True Negative, False Positive, and False Negative, respectively.
Thus, our proposed method generated inaccurate predictions for a total of 196 samples. From these failure cases, we randomly selected 100 samples and performed a detailed manual analysis of the errors. Table~\ref{tab:error} summarizes the failure modes observed in these cases. We identified five primary reasons for the prediction failures:

\textbf{Object Grounding Error:} 
This refers to cases in which the model’s grounding capability was inadequate, thus the model failed to recognize which specific object was being operated on.

\textbf{Action Comprehension Error:}
These instances arise when the model misinterprets the possible trajectories and interactions described by $\bm{x}_\mathrm{txt}$. 
For example, while $\bm{x}_\mathrm{txt}$ is ``Open the drawer,'' the manipulator attempts to interact with an object on top of the drawer instead, it is clearly an error. 
This category includes cases where the model makes incorrect predictions despite the fact that understanding at least the verb would indicate the mistake.

\textbf{Referring Expression Comprehension Error:}
As in Fig.~\ref{fig:qualitative} (iii), this pertains to cases in which $\bm{x}_\mathrm{txt}$ contains referring expressions (such as `next to the cola can' or `near the plastic bottle'), and the model fails to properly interpret these expressions.

\textbf{Partial Visibility:}
This category involves scenarios where the target object or location is only partially visible, which complicates accurate prediction. Such situations may arise if more than half of the target is obscured by the manipulator or other objects, or if the majority of the target is positioned outside the camera’s field of view. It also includes cases where the end-effector was not visible in $\bm{x}_\mathrm{img}$.

\textbf{Annotation Error:}
This covers cases in which the ground truth label appears to have been assigned incorrectly.

According to Table~\ref{tab:error}, Object Grounding Errors were particularly detrimental. 
We attribute this to the absence of a specialized component within our method for detecting each object and grounding the language representations in the instructions to those objects.
This indicates that there is room to improve the text-image alignment. 
Potential solutions include using a multimodal large language model to describe every object in the image, thus achieving better alignment between text and image representations, or implementing this mechanism via an open-vocabulary object detector (e.g., Detic~\cite{zhou2022detecting}).

\section{Conclusions}

In this study, we focused on verifying the pre-manipulation alignment between a natural language instruction, a pre-manipulation egocentric image, and a planned end-effector trajectory.

Our contributions are as follows:
\begin{itemize}
\item We proposed a model that predicts whether the task specified by an instruction sentence can be appropriately executed by aligning the trajectory with the pre-manipulation image.
\item We introduced the MTF module, which employs a state-of-the-art state-space model, S4, and the transformer encoder in parallel to capture multi-level time-series self-correlation within the end effector trajectory.
\item We also introduced the Trimodal Cross-Integration module to effectively align natural language instructions, camera images, and the end effector trajectories.
\item Experimental results demonstrated that our method achieved higher prediction accuracy than several baseline methods.
\end{itemize}

As a potential future direction, it is conceivable to combine the proposed approach with VLA models~\cite{octo_2023, brohan2022rt, kim24openvla}. 
As discussed in Sec.~\ref{intro}, many current VLA models exhibit insufficient alignment between modalities, which can lead to the generation of inappropriate trajectories.
Moreover, in most cases, those models lack mechanisms to evaluate the validity of the trajectories they generate, making it challenging to prevent the execution of erroneous trajectories. 
By applying the proposed method to these methods, it becomes possible to select and execute the suitable trajectory from a set of candidates generated by VLA models, thereby enabling more effective execution of object manipulation tasks.

While our study focuses on table-top manipulation, we note that incorporating physical information, such as force and affordance, could potentially improve the accuracy of prediction in such settings. 
For example, failures in grasping tasks can often be readily detected using force sensors. 
In addition, attempts to interact with objects beyond the reachable workspace of the robot can be preemptively identified as infeasible considering the robot's accessibility. 
These observations suggest that the integration of physical information into model inputs is a promising direction for future research.

\vspace{-2mm}
\section*{ACKNOWLEDGMENT}
This work was partially supported by JSPS KAKENHI Grant Number 23H03478, JST Moonshot, and JSPS Fellows Grant Number JP23KJ1917.
\vspace{-1mm}
\bibliographystyle{tfnlm}
\bibliography{reference}

\begin{thebibliography}{10}
\providecommand{\url}[1]{\normalfont{#1}}
\providecommand{\urlprefix}{Available from: }

\bibitem{wu2023tidybot}
Wu~J, Antonova~R, Kan~A, et~al. {Tidybot: Personalized Robot Assistance with Large Language Models}. Autonomous Robots. 2023;\hspace{0pt}47(8):1087--1102.

\bibitem{lehnert2017autonomous}
Lehnert~C, English~A, McCool~C, et~al. {Autonomous Sweet Pepper Harvesting for Protected Cropping Systems}. IEEE RA-L. 2017;\hspace{0pt}2(2):872--879.

\bibitem{jun2021towards}
Jun~J, Kim~J, Seol~J, et~al. {Towards an Efficient Tomato Harvesting Robot: 3D Perception, Manipulation, and End-Effector}. IEEE Access. 2021;\hspace{0pt}9:17631--17640.

\bibitem{goko2024task}
Goko~M, Kambara~M, Saito~D, et~al. {Task Success Prediction for Open-Vocabulary Manipulation Based on Multi-Level Aligned Representations}. In: CoRL; 2024.

\bibitem{xiao2022skill}
Xiao~T, Chan~H, Sermanet~P, et~al. {Skill Acquisition by Instruction Augmentation on Offline Datasets}. In: LangRob @ CoRL22; 2022.

\bibitem{liu2023reflect}
Liu~Z, Bahety~A, Song~S. {REFLECT: Summarizing Robot Experiences for Failure Explanation and Correction}. In: CoRL; 2023. p. 3468--3484.

\bibitem{inceoglu2021fino}
Inceoglu~A, Aksoy~EE, Ak~AC, et~al. {Fino-Net: A Deep Multimodal Sensor Fusion Framework for Manipulation Failure Detection}. In: IROS; 2021. p. 6841--6847.

\bibitem{driess2023palm}
Driess~D, Xia~F, Sajjadi~MS, et~al. {PaLM-E: An Embodied Multimodal Language Model}. In: ICML; 2023. p. 8469--8488.

\bibitem{diehl2022did}
Diehl~M, Ramirez-Amaro~K. {Why Did I Fail? A Causal-Based Method to Find Explanations for Robot Failures}. IEEE Robotics and Automation Letters. 2022;\hspace{0pt}7(4):8925--8932.

\bibitem{das2021semantic}
Das~D, Chernova~S. {Semantic-Based Explainable AI: Leveraging Semantic Scene Graphs and Pairwise Ranking to Explain Robot Failures}. In: IROS; 2021. p. 3034--3041.

\bibitem{octo_2023}
{Octo Model Team}, Ghosh~D, Walke~H, et~al. {Octo: An Open-Source Generalist Robot Policy}. In: RSS; 2024.

\bibitem{brohan2022rt}
Brohan~A, Brown~N, Carbajal~J, et~al. {RT-1: Robotics Transformer for Real-World Control at Scale}. arXiv preprint arXiv:221206817. 2022;\hspace{0pt}.

\bibitem{kim24openvla}
Kim~M, Pertsch~K, Karamcheti~S, et~al. {OpenVLA: An Open-Source Vision-Language-Action Model}. In: CoRL; 2024.

\bibitem{black2024pi_0}
Black~K, Brown~N, Driess~D, et~al. {$\pi_0$: A Vision-Language-Action Flow Model for General Robot Control}. arXiv preprint arXiv:241024164. 2024;\hspace{0pt}.

\bibitem{vaswani2017attention}
Vaswani~A, Shazeer~N, Parmar~N, et~al. {Attention Is All You Need}. In: NIPS; 2017.

\bibitem{s4}
Gu~A, Goel~K, R{\'{e}}~C. {Efficiently Modeling Long Sequences with Structured State Spaces}. In: ICLR; 2022.

\bibitem{haddadin2017robot}
Haddadin~S, Luca~A, Albu-Schäffer~A. {Robot Collisions: A Survey on Detection, Isolation, and Identification}. T-RO. 2017;\hspace{0pt}33(6):1292--1312.

\bibitem{mottaghi2016what}
Mottaghi~R, Rastegari~M, Gupta~A, et~al. {``What Happens If...'' Learning to Predict the Effect of Forces in Images}. In: ECCV; 2016. p. 269--285.

\bibitem{magassouba2021predicting}
Magassouba~A, Sugiura~K, Nakayama~A, et~al. {Predicting and Attending to Damaging Collisions for Placing Everyday Objects in Photo-Realistic Simulations}. Advanced Robotics. 2021;\hspace{0pt}35(12):787--799.

\bibitem{kambara2022relational}
Kambara~M, Sugiura~K. {Relational Future Captioning Model for Explaining Likely Collisions in Daily Tasks}. In: ICIP; 2022. p. 2601--2605.

\bibitem{liu2024model}
Liu~H, Dass~S, Mart{\'\i}n-Mart{\'\i}n~R, et~al. {Model-Based Runtime Monitoring with Interactive Imitation Learning}. In: ICRA; 2024. p. 4154--4161.

\bibitem{kawaharazuka2024real}
Kawaharazuka~K, Matsushima~T, Gambardella~A, et~al. {Real-World Robot Applications of Foundation Models: A Review}. arXiv preprint arXiv:240205741. 2024;\hspace{0pt}.

\bibitem{brohan2023saycan}
Brohan~A, Chebotar~Y, Finn~C, et~al. {Do As I Can, Not As I Say: Grounding Language in Robotic Affordances}. In: CoRL; 2023. p. 287--318.

\bibitem{shirasaka2024selfrecovery}
Shirasaka~M, Matsushima~T, Tsunashima~S, et~al. {Self-Recovery Prompting: Promptable General Purpose Service Robot System with Foundation Models and Self-Recovery}. In: ICRA; 2024.

\bibitem{Driess2023palme}
Driess~D, Xia~F, Sajjadi~M, et~al. {PaLM-E: An Embodied Multimodal Language Model}. In: ICML; 2023. p. 8469--8488.

\bibitem{zha2024distilling}
Zha~L, Cui~Y, Lin~LH, et~al. {Distilling and Retrieving Generalizable Knowledge for Robot Manipulation via Language Corrections}. In: ICRA; 2024. p. 15172--15179.

\bibitem{sun2023interactive}
Sun~L, Jha~D, Hori~C, et~al. {Interactive Planning Using Large Language Models for Partially Observable Robotics Tasks}. In: ICRA; 2024.

\bibitem{zhang2024feedback}
Zhang~J, Huang~Z, Ray~A, et~al. {Feedback-Guided Autonomous Driving}. In: CVPR; 2024. p. 15000--15011.

\bibitem{zhi2024closed}
Zhi~P, Zhang~Z, Han~M, et~al. {Closed-Loop Open-Vocabulary Mobile Manipulation with GPT-4V}. arXiv preprint arXiv:240410220. 2024;\hspace{0pt}.

\bibitem{lin2024drplanner}
Lin~Y, Li~C, Ding~M, et~al. {DrPlanner: Diagnosis and Repair of Motion Planners for Automated Vehicles using Large Language Models}. IEEE RA-L. 2024;\hspace{0pt}.

\bibitem{xiong2024autonomous}
Xiong~C, Shen~C, Li~X, et~al. {Autonomous Interactive Correction MLLM for Robust Robotic Manipulation}. In: CoRL; 2024.

\bibitem{sermanet2024robovqa}
Sermanet~P, Ding~T, Zhao~J, et~al. {RoboVQA: Multimodal Long-Horizon Reasoning for Robotics}. In: ICRA; 2024. p. 645--652.

\bibitem{duan24aha}
Duan~J, Pumacay~W, Kumar~N, et~al. {AHA: A Vision-Language-Model for Detecting and Reasoning Over Failures in Robotic Manipulation}. In: 1st Workshop on X-Embodiment Robot Learning; 2024.

\bibitem{liu2023robofail}
Liu~Z, Bahety~A, Song~S. {REFLECT: Summarizing Robot Experiences for Failure Explanation and Correction}. In: CoRL; 2023. p. 3468--3484.

\bibitem{antol2015vqa}
Antol~S, Agrawal~A, Lu~J, et~al. {VQA: Visual Question Answering}. In: ICCV; 2015. p. 2425--2433.

\bibitem{yin2023survey}
Yin~S, Fu~C, Zhao~S, et~al. {A Survey on Multimodal Large Language Models}. arXiv preprint arXiv:230613549. 2023;\hspace{0pt}.

\bibitem{wu2017visual}
Wu~Q, Teney~D, Wang~P, et~al. {Visual Question Answering: A Survey of Methods and Datasets}. Computer Vision and Image Understanding. 2017;\hspace{0pt}163:21--40.

\bibitem{wang2024qwen2}
Wang~P, Bai~S, Tan~S, et~al. {Qwen2-VL: Enhancing Vision-Language Model's Perception of The World at Any Resolution}. arXiv preprint arXiv:240912191. 2024;\hspace{0pt}.

\bibitem{li2023blip}
Li~J, Li~D, Savarese~S, et~al. {BLIP-2: Bootstrapping Language-Image Pre-Training with Frozen Image Encoders and Large Language Models}. In: ICML; 2023. p. 19730--19742.

\bibitem{bai2023qwen}
Bai~J, Bai~S, Yang~S, et~al. {Qwen-VL: A Frontier Large Vision-Language Model with Versatile Abilities}. arXiv preprint arXiv:230812966. 2023;\hspace{0pt}.

\bibitem{wang2023image}
Wang~W, Bao~H, Dong~L, et~al. {Image as A Foreign Language: Beit Pretraining for Vision and Vision-Language Tasks}. In: CVPR; 2023. p. 19175--19186.

\bibitem{liu2024improved}
Liu~H, Li~C, Li~Y, et~al. {Improved Baselines with Visual Instruction Tuning}. In: CVPR; 2024. p. 26296--26306.

\bibitem{liu2024visual}
Liu~H, Li~C, Wu~Q, et~al. {Visual Instruction Tuning}. In: NeurIPS; 2023. p. 34892--34916.

\bibitem{instructblip}
Dai~W, Li~J, Li~D, et~al. {InstructBLIP: Towards General-purpose Vision-Language Models with Instruction Tuning}. In: NeurIPS; 2023.

\bibitem{hudson2019gqa}
Hudson~D, Manning~C. {GQA: A New Dataset for Real-World Visual Reasoning and Compositional Question Answering}. In: CVPR; 2019. p. 6700--6709.

\bibitem{goyal2017making}
Goyal~Y, Khot~T, Summers-Stay~D, et~al. {Making The V in VQA Matter: Elevating The Role of Image Understanding in Visual Question Answering}. In: CVPR; 2017. p. 6904--6913.

\bibitem{lssl}
Gu~A, et~al. {Combining Recurrent, Convolutional, and Continuous-time Models with Linear State-Space Layers}. In: NeurIPS; Vol.~34; 2021.

\bibitem{s4d}
Gu~A, Gupta~A, Goel~K, et~al. {On the Parameterization and Initialization of Diagonal State Space Models}. In: NeurIPS; 2022. p. 35971--35983.

\bibitem{s5}
Smith~JTH, Warrington~A, Linderman~SW. {Simplified State Space Layers for Sequence Modeling}. In: ICLR; 2023.

\bibitem{classic_ssm}
Funahashi~Ki, Nakamura~Y. {Approximation of Dynamical Systems by Continuous Time Recurrent Neural Networks}. Neural networks. 1993;\hspace{0pt}6(6):801--806.

\bibitem{classic_ssm2}
Tallec~C, Ollivier~Y. {Can Recurrent Neural Networks Warp Time?} In: ICLR; 2018.

\bibitem{gu2024mamba}
Gu~A, Dao~T. {Mamba: Linear-Time Sequence Modeling with Selective State Spaces}. In: CoLM; 2024.

\bibitem{zhuvision}
Zhu~L, Liao~B, Zhang~Q, et~al. {Vision Mamba: Efficient Visual Representation Learning with Bidirectional State Space Model}. In: ICML; 2024.

\bibitem{zoh}
Zhang~Z, et~al. {Comparison Between First-order Hold With Zero-order Hold in Discretization of Input-delay Nonlinear Systems}. In: ICCAS; 2007. p. 2892--2896.

\bibitem{stellaen}
{Stella en 1.5B v5} ; 2024. {Accessed: Dec. 2024}; \urlprefix\url{https://huggingface.co/dunzhang/stella_en_1.5B_v5}.

\bibitem{oquab2024dinov2}
Oquab~M, Darcet~T, Moutakanni~T, et~al. {DINOv2: Learning Robust Visual Features without Supervision}. TMLR. 2024;\hspace{0pt}:1--31.

\bibitem{gpt4o}
{OpenAI}. {GPT-4o} ; {Accessed: Nov. 2024.} \urlprefix\url{https://platform.openai.com/docs/models/gpt-4o}.

\bibitem{kuangram}
Kuang~Y, Ye~J, Geng~H, et~al. {RAM: Retrieval-Based Affordance Transfer for Generalizable Zero-Shot Robotic Manipulation}. In: CoRL; 2024.

\bibitem{bahl2023affordances}
Bahl~S, Mendonca~R, Chen~L, et~al. {Affordances from Human Videos as A Versatile Representation forRrobotics}. In: CVPR; 2023. p. 13778--13790.

\bibitem{hwang2025motif}
Hwang~M, Hejna~J, Sadigh~D, et~al. {MotIF: Motion Instruction Fine-tuning}. IEEE RA-L. 2025;\hspace{0pt}.

\bibitem{zhou2022detecting}
Zhou~X, Girdhar~R, Joulin~A, et~al. {Detecting Twenty-Thousand Classes using Image-Level Supervision}. In: ECCV; 2022. p. 350--368.

\end{thebibliography}
\end{document}